\documentclass[11pt]{article}

\usepackage[margin=1in]{geometry}
\usepackage{amsmath,amssymb,amsthm,amsfonts}
\usepackage{amsopn}
\usepackage{graphicx}
\usepackage{epstopdf}
\usepackage{siunitx}
\usepackage{mathtools}      
\usepackage{nicefrac}       
\usepackage[expansion=false]{microtype}      
\usepackage{xcolor}
\usepackage{multirow}
\usepackage{booktabs}       
\usepackage{algorithm}
\usepackage{algpseudocode}
\usepackage{float}          
\usepackage{makecell}
\usepackage{bbold}
\usepackage{hyperref}

\ifpdf
  \DeclareGraphicsExtensions{.eps,.pdf,.png,.jpg}
\else
  \DeclareGraphicsExtensions{.eps}
\fi

\newtheorem{proposition}{Proposition}
\newtheorem{definition}{Definition}

\DeclareMathOperator*{\argmax}{argmax}

\newcommand{\R}{\mathbb{R}}

\newcommand{\bs}{\boldsymbol}

\title{Multiclass threshold-based classification and model evaluation}

\author{
Edoardo Legnaro\\
Department of Mathematics (DIMA), University of Genova, Genova, Italy\\
\texttt{edoardo.legnaro@edu.unige.it}
\and
Sabrina Guastavino\\
Department of Mathematics (DIMA), University of Genova, Genova, Italy\\
\texttt{sabrina.guastavino@unige.it}
\and
Francesco Marchetti\\
Department of Mathematics ``Tullio Levi-Civita'', University of Padova, Padova, Italy\\
\texttt{francesco.marchetti@unipd.it}
}

\hypersetup{
  pdftitle={Multiclass threshold-based classification and model evaluation},
  pdfauthor={E. Legnaro, S. Guastavino, and F. Marchetti},
  colorlinks=true,
  linkcolor=blue,
  citecolor=blue,
  urlcolor=blue
}

\begin{document}

\maketitle

\begin{abstract}
In this paper, we introduce a threshold-based framework for multiclass classification that generalizes the standard argmax rule. This is done by replacing the probabilistic interpretation of softmax outputs with a geometric one on the multidimensional simplex, where the classification depends on a multidimensional threshold. This change of perspective enables for any trained classification network an \textit{a posteriori} optimization of the classification score by means of threshold tuning, as usually carried out in the binary setting, thus allowing for a further refinement of the prediction capability of any network. 
Our experiments show indeed that multidimensional threshold tuning yields performance improvements across various networks and datasets. Moreover, we derive a multiclass ROC analysis based on \emph{ROC clouds}---the attainable (FPR,TPR) operating points induced by a single multiclass threshold---and summarize them via a \emph{Distance From Point} (DFP) score to $(0,1)$. This yields a coherent alternative to standard One-vs-Rest (OvR) curves and aligns with the observed tuning gains.
\end{abstract}



\section{Introduction}

In the standard supervised classification setting, a fundamental distinction is made between the binary and multiclass framework. Although some machine learning methods are natively binary, in deep learning this distinction leads to two different types of output of the neural network, which are handled differently in order to get predictions and then performance assessment \cite{Goodfellow16,zhang2000neural}. 

\subsection{The binary case}

In the binary classification setting, each element $\bs{x}_i$ of the $d$-dimensional dataset $\mathcal{X}=\{\bs{x}_1,\dots,\bs{x}_n\}\subset\Omega\subset\mathbb{R}^d$ is associated with a label $y_i\in\{0,1\}$. The aim is to learn the sample-label association by means of a network that produces the output
\begin{equation*}
   \hat{y}_{\bs{\theta}}(\bs{x})=(\sigma\circ  h)(\bs{x},\bs{\theta})\in [0,1],\; \bs{x}\in\Omega,
\end{equation*}
where $h(\bs{x},\bs{\theta})$ denotes the outcome of the \textit{input} and \textit{hidden} layers, which depend on a vector (matrix) of weight parameters $\bs{\theta}$, and $\sigma$ is the well-known \textit{sigmoid} activation function defined as $\sigma(h)=(1+e^{-h})^{-1}$. 

Then, the classification of a sample $\bs{x}$ to one class or the other depends on a threshold, that is, the corresponding output $\hat{y}=\hat{y}_{\bs{\theta}}(\bs{x})$ is assigned to one of the two labels $\{y=0\}$ or $\{y=1\}$ by means of a function:
\begin{equation*}
	\mathbb{1}_{\hat{y}}(\tau)=\mathbb{1}_{\{\hat{y}>\tau\}}=\begin{dcases} 0 & \textrm{if } \hat{y}\le \tau,\\
		1 & \textrm{if }\hat{y}>\tau,\end{dcases}
\end{equation*}
being $\tau\in(0,1)$ a threshold value. 

Once the network is trained by minimizing some objective function, the performance of the model is usually assessed by means of some classification score $s$, which is a function that takes input from the entries of the confusion matrix:
\begin{equation}\label{eq:cm}
	\mathrm{CM}(\tau,\bs{\theta}) =
	\begin{pmatrix}
		\mathrm{TN}(\tau,\bs{\theta}) & \mathrm{FP}(\tau,\bs{\theta}) \\
		\mathrm{FN}(\tau,\bs{\theta}) & \mathrm{TP}(\tau,\bs{\theta})
	\end{pmatrix},
\end{equation}
being
\begin{equation}\label{eq:mat_defi}
	\begin{split}
		& \mathrm{TN}(\tau,\bs{\theta}) = \sum_{i=1}^n{(1-y_i)\mathbb{1}_{\{\hat{y}_{\bs{\theta}}(\bs{x}_i)<\tau\}}}, \quad \mathrm{TP}(\tau,\bs{\theta}) = \sum_{i=1}^n{y_i\mathbb{1}_{\{\hat{y}_{\bs{\theta}}(\bs{x}_i)>\tau\}}},\\
		& \mathrm{FP}(\tau,\bs{\theta}) = \sum_{i=1}^n{(1-y_i)\mathbb{1}_{\{\hat{y}_{\bs{\theta}}(\bs{x}_i)>\tau\}}},\quad \mathrm{FN}(\tau,\bs{\theta}) = \sum_{i=1}^n{y_i\mathbb{1}_{\{\hat{y}_{\bs{\theta}}(\bs{x}_i)<\tau\}}}.
	\end{split}
\end{equation}
We note that a meaningful score is not decreasing with respect to $\mathrm{TN}$ and $\mathrm{TP}$ and is not increasing with respect to $\mathrm{FN}$ and $\mathrm{FP}$. The reliance on a threshold has allowed the development of a straightforward solution for \textit{a posteriori} score optimization, that is, once the network is trained and the \textit{optimal} weights $\bs{\theta}^{\star}$ are defined, find the threshold value that maximizes your classification metric of interest
\begin{equation}\label{eq:a_posteriori}
	\max_{\tau\in(0,1)}s(\mathrm{CM}(\tau,\bs{\theta}^\star)).
\end{equation}
This tuning, which is performed using the training or validation set, can severely improve the performance of the network, especially when settings with unbalanced classes are considered, and it represents a common practice in this framework.
We also recall that, in the binary setting, \emph{score-oriented loss} functions have been proposed to incorporate the desired evaluation metric directly into the training objective via probabilistic confusion matrices \cite{Marchetti22}.

In addition or alternatively to threshold tuning, classifier performance can be evaluated across all possible thresholds, which is the main principle underpinning the well-known Receiver Operating Characteristic (ROC) analysis. In particular, the ROC curve plots the true positive rate (TPR, or sensitivity) against the false positive rate (FPR, or 1-specificity) as the decision threshold is varied \cite{fawcett2006introduction}. Here, TPR is defined as the proportion of correctly identified positive samples, while FPR is the proportion of negative samples incorrectly classified as positive. This graphical approach provides a comprehensive view of a model's discriminative ability, independent of any specific threshold choice. To summarize the information provided by the ROC curve, the area under the curve (AUC) is often reported and can be interpreted as the probability that a randomly chosen positive instance is ranked higher than a randomly chosen negative instance \cite{bradley1997use}. Precisely, an AUC of 1 indicates perfect classification, while an AUC of 0.5 corresponds to random guessing, which corresponds to the case where the ROC curve is the diagonal bisector line. 
ROC and AUC are thus standard tools for comparing classifiers, especially in imbalanced or cost-sensitive scenarios.

\subsection{The multiclass case}
In contrast to the binary setting, when approaching a multiclass problem where $C_1,\dots,C_m$ are $m>2$ classes and each $y_i\in C_j$ for a unique $j=1,\dots,m$, the classification is no longer based on a threshold value. Instead, $\sigma$ is a \textit{softmax} activation function, which models the output as
\begin{equation*}
    \hat{\bs{y}}_{\bs{\theta}} = (\hat{y}_{\bs{\theta}}^1,\dots,\hat{y}_{\bs{\theta}}^m),\quad \sum_{j=1}^m \hat{y}_{\bs{\theta}}^j = 1.
\end{equation*}
Then, a sample $\bs{x}$ is usually assigned to the class $j^\star$ that represents the \textit{argmax} of $\hat{\bs{y}}_{\bs{\theta}}$ with respect to the $m$ classes, that is, the output is interpreted as a probability distribution and \cite{murphy2018machine}
\begin{equation*}
	j^\star = \textrm{argmax}_{j=1,\dots,m}\hat{y}_{\bs{\theta}}^j.
\end{equation*}
Although a single sigmoid unit is usually chosen because of its simpler implementation and interpretation, we point out that two softmax units can be used in the binary framework. In this case, the value of one unit is redundant, and the argmax rule plays the role of a 0.5 threshold value.

Alternative prediction rules to the standard argmax have been explored in several works on multiclass classification \cite{bridle1990probabilistic,martins2016softmax}. For example, in \cite{roberts2023geometry} the replacement of the argmax with the Fréchet mean was proposed, which enables pre-trained models to generalize to novel classes by leveraging the geometric structure of the label space, without requiring additional training. More recently, \cite{soloff2024building} introduced a stable relaxation of the argmax, termed the ``inflated argmax'', with the objective of improving the stability of multiclass classifiers.

In the multiclass scenario, the classification results are very often evaluated in terms of an extension of the binary score to the multiclass case, which is computed by considering $m$ \textit{one-vs-rest} (OvR) confusion matrices \eqref{eq:cm}, where a class is considered positive and the set of remaining ones as negative. However, in this manner, a posteriori score optimization cannot be performed via some threshold variation if an argmax-like classification rule is applied, or at least there is no native way to do so. To obtain some a posteriori improvement in the classification performance, another strategy is represented by the so-called model calibration, whose goal is to postprocess a model's scores so that they better align with true class probabilities. This is typically achieved by training a secondary model, such as a Platt scaler or an isotonic regressor, on the outputs of a pre-trained classifier \cite{guo2017calibration}. In fact, since they deal with OvR-computed scores, well-established standard multiclass calibration techniques are extensions of binary methods and often rely on an OvR decomposition \cite{zadrozny2002transforming}, where a separate calibrator is trained for each class, thus losing the original intertwined structure of the multiclass network output. Analogous issues affect ROC analysis in the presence of more than two classes. In fact, the OvR approach yields to multiple ROC curves, one for each class \cite{hand2001simple}, which are then aggregated using macro-averaged or weighted-averaged AUC scores. However, simply averaging summarizing scalar metrics could lead to oversimplification, and, in a sense that will be deepened in \ref{sec:roc}, such a procedure relies on decision rules that are \textit{unnatural} for the multiclass framework.
These approaches fall largely into two main categories: those that extend decomposition strategies and those that attempt a more direct geometric generalization. Another influential decomposition strategy is the One-vs-One (OvO) approach, which trains a unique binary classifier for every pair of classes \cite{allwein2000reducing, wandishin2009multiclass}. 
For a problem with $K$ classes, this results in $K(K-1)/2$ classifiers \cite{wandishin2009multiclass}. In \cite{hand2001simple} a multiclass AUC measure was proposed, now commonly known as MAUC, based on this principle by averaging the AUCs of all pairwise comparisons. The second major branch of research pursues a more theoretically direct generalization by defining a high-dimensional ROC ``hypersurface'' \cite{holmes2002multiclass, wandishin2009multiclass}. In this framework, the performance of a classifier is summarized by Volume Under the Surface (VUS) or Hypervolume Under the Manifold (HUM) \cite{ferri2003volume, landgrebe2008efficient, kleiman2019aucmu}. 

\subsection{Our contribution}

In this paper, our aim is to provide a novel framework by introducing a threshold-based setting for the multiclass case that generalizes the classical argmax operation. This is obtained by discarding the probabilistic interpretation of the softmax-based output and considering it in its natural domain, which is the multidimensional simplex \cite{tang2019novel}. We observe that simplex geometry has previously been used for multiclass classification in various approaches for different purposes. In \cite{mroueh2012multiclass}, the classification task is reformulated as a vector-valued regression problem, where the model learns to map inputs to points in a regular simplex, allowing for a geometrically structured representation of class labels. In \cite{heese2023calibrated}, the simplex is instead utilized in the latent space: the training data is embedded in a space whose geometry is defined by a regular $(m-1)$-dimensional simplex, with $m$ denoting the number of classes. 

By considering the output vector as a single point in the simplex, our method offers a natively multiclass framework that preserves the relationships between all class scores simultaneously. The novelty of our approach lies in performing classification by relating each class to a corresponding subset of the simplex, thanks to the introduction of a multidimensional threshold parameter, which allows for
\begin{itemize}
    \item 
    a native a posteriori optimization of a score, which can be easily applied to any trained network, as in the binary setting;
    \item 
    a consequent \textit{coherent} ROC analysis, which is aligned with the multiclass network output, differently with respect to the standard strategies such as OvR or OvO.
\end{itemize}
We point out that our approach, despite operating on the model's output, is conceptually distinct from calibration methods, being an inference-time decision rule that operates directly on the raw, uncalibrated output vector provided by the softmax function, and requires no additional training or post-hoc model fitting.

Building on the simplex decision rule, we introduce \emph{ROC clouds} for multiclass models: for each multidimensional threshold in the simplex, we obtain a coherent, natively multiclass operating point that yields classwise $(\mathrm{FPR},\mathrm{TPR})$ pairs. 
Varying the threshold generates, for each class, a point cloud in the ROC plane that reflects global, joint trade-offs under a single decision rule. We summarize each cloud with a \emph{Distance From Point} (DFP) score---the mean $L_1$ distance to $(0,1)$ over sampled thresholds---which provides a compact, threshold-agnostic evaluation aligned with our tuning framework.

The paper is organized as follows. In \ref{sec:multi_thresh}, we formalize our novel threshold-based framework, from which we derive the threshold tuning algorithm in \ref{sec:score_opt}. This allows for a linked multiclass ROC analysis, which is described in detail and compared to the standard approach in \ref{sec:roc}. The results presented in \ref{sec:results} show that multidimensional threshold tuning is effective in improving the classification performance obtained by means of the classical argmax rule. 

\section{Predicting in the multidimensional simplex}\label{sec:multi_thresh}

In this section, our purpose is to show that a threshold-based framework can be recovered in the multiclass setting too, allowing an improved flexible handling of the output with the aim of then enhancing score performance and model analysis. To provide a clear setting and subsequent, we will explicitly formalize some crucial concepts.

First, we observe that the output $\hat{\bs{y}}$ is contained in the $(m-1)$-simplex $S_m=\{\bs{z}\in\R^m\:|\: \sum_{j=1}^m \bs{z}^j = 1 \}$, whose vertices are the \textit{one-hot} encoded classes $\bs{e}_1,\dots,\bs{e}_m$ corresponding to $C_1,\dots,C_m$ \cite{Harris13}. Then, we provide the following definition.
\begin{definition}[Simplex classification collection]\label{def:region}
Let $\bs{\tau}=(\tau^1,\dots,\tau^m)\in S_m$ be a multidimensional threshold. A simplex classification collection $R_1(\bs{\tau}),\dots,R_m(\bs{\tau})$ of classification regions $R_j(\bs{\tau})$ for the class $C_j$, $j=1,\dots,m$, satisfies the following properties.
\begin{enumerate}
    \item
     $R_j(\bs{\tau})\subset S_m$ for $j=1,\dots,m$,
     \item 
     $\bs{e}_j\in R_j(\bs{\tau})$ for $j=1,\dots,m$,
     \item
     $\textrm{cl}\big(\bigcup_{j=1}^m{R_j(\bs{\tau})}\big)=S_m$,
\end{enumerate}
being $\textrm{cl}$ the topological closure.
\end{definition}
According to \ref{def:region}, any point in the simplex is assigned to at least one classification region $R_1(\bs{\tau}),\dots,R_m(\bs{\tau})$, but a set of points that lie in subsets of $S_m$ of null measure. This ambiguity is actually present in classical binary and multiclass classification. For example, in an argmax-driven $3$-class problem, the output $(0.5,0.5,0)$ is in principle undecided between the first and the second class. 

Another critical point in \ref{def:region} is the fact that one output in the simplex could be assigned to more than one class. In \ref{sec:roc}, we will discuss in which way this is connected to classical ROC analysis, but with the following definition we restrict ourselves to simplex classification collections that are not affected by this multilabel temptation.
\begin{definition}[Proper simplex classification collection]\label{def:region_proper}
Let $\bs{\tau}=(\tau^1,\dots,\tau^m)\in S_m$ be a multidimensional threshold. A simplex classification collection $R_1(\bs{\tau}),\dots,R_m(\bs{\tau})$ is called proper if
\begin{equation*}
     R_k(\bs{\tau})\cap R_j(\bs{\tau})=\emptyset 
\end{equation*}
for $k,j=1,\dots,m$, $k\neq j$.
\end{definition}
Therefore, a proper classification collection consists of a sort of partition of the simplex $S_m$, where each output is now assigned to one class, besides null measure subsets, which is a crucial aspect to get a well-posed classification scheme. In particular, note that varying $\bs{\tau}$ will produce a variation in the corresponding \textit{classification rule}, i.e., the procedure that assigns a class to the received output.

\begin{figure}[htbp]
	\centering
	\includegraphics[width=\linewidth]{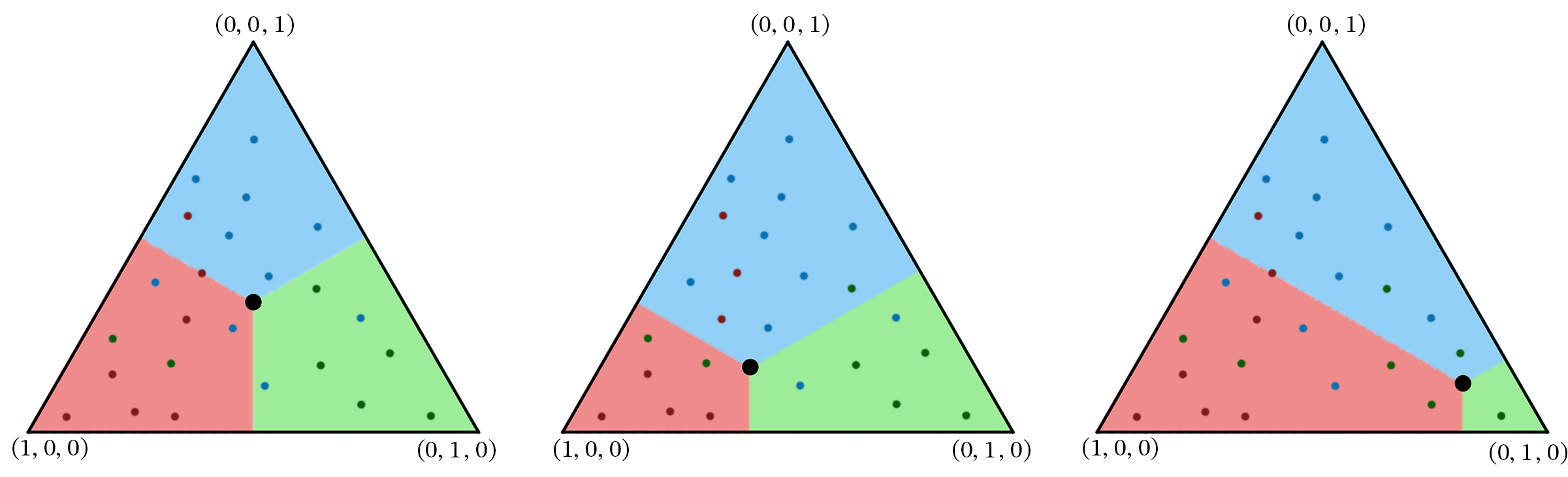} 
	\caption{For $m=3$, the three regions $R_1(\bs{\tau})$ (red), $R_2(\bs{\tau})$ (green) and $R_3(\bs{\tau})$ (blue). From left to right: $\bs{\tau}=(1/3,1/3,1/3)$, $\bs{\tau}=(1/2,1/3,1/6)$, $\bs{\tau}=(1/8,3/4,1/8)$ (big black dot). The color blue, red or green represents the true label of the samples (colored dots). Number of misclassifications from left to right: $7$, $8$ and $10$.}
	\label{fig:simplices}
\end{figure}

Among the possible classification collections that can be defined in the simplex, in the following we consider classification regions defined as
\begin{equation}\label{eq:natural}
    R_j(\bs{\tau})=\{\bs{z}\in S_m\:|\: z^j-z^k>\tau^j-\tau^k,\; k\neq j\},\; j=1,\dots,m.
\end{equation}
We will refer to this choice as the \textit{natural} classification collection, as we motivate in the following proposition.
\begin{proposition}
    The simplex classification collection defined in \eqref{eq:natural} is proper and generalizes the argmax procedure, i.e., the classical argmax is a particular case of \eqref{eq:natural}.
\end{proposition}
\begin{proof}
    The collection is trivially proper, because $\bs{z}\in R_k(\bs{\tau})\cap R_j(\bs{\tau})$, $k\neq j$, would imply both $z^j>z^k$ and $z^k>z^j$. Then, if $\tau^j=1/m$ for any $j=1,\dots,m$, then $R_j(\bs{\tau})=\{\bs{z}\in S_m\:|\: z^j>z^k,\; k\neq j\}$ leads to the classical argmax procedure.
\end{proof}
See \ref{fig:simplices} for examples of natural classification collections obtained by varying the threshold vector.

\section{Multiclass threshold-based optimization}\label{sec:score_opt}

The threshold-based framework introduced in the previous section can be used to obtain an \textit{optimized} multiclass classification rule by performing a posteriori threshold tuning, analogously to the binary case.

Letting $s$ be a binary score, for each $j=1,\dots,m$ we can consider
\begin{equation*}
\mathrm{CM}_j(\bs{\tau},\bs{\theta}) =
\begin{pmatrix}
\mathrm{TN}_j(\bs{\tau},\bs{\theta}) & \mathrm{FP}_j(\bs{\tau},\bs{\theta}) \\
\mathrm{FN}_j(\bs{\tau},\bs{\theta}) & \mathrm{TP}_j(\bs{\tau},\bs{\theta})
\end{pmatrix},
\end{equation*}
with
\begin{equation*}
    \begin{split}
         & \mathrm{TN}_j(\bs{\tau},\bs{\theta}) = \sum_{i=1}^n{\mathbb{1}_{\{\bs{y}_i\neq \bs{e}_j\}}\mathbb{1}_{\{\hat{\bs{y}}_{\bs{\theta}}(\bs{x}_i)\notin R_j(\bs{\tau})\}}},\; \mathrm{TP}_j(\bs{\tau},\bs{\theta}) = \sum_{i=1}^n{\mathbb{1}_{\{\bs{y}_i= \bs{e}_j\}}\mathbb{1}_{\{\hat{\bs{y}}_{\bs{\theta}}(\bs{x}_i)\in R_j(\bs{\tau})\}}},\\
         & \mathrm{FP}_j(\bs{\tau},\bs{\theta}) = \sum_{i=1}^n{\mathbb{1}_{\{\bs{y}_i\neq \bs{e}_j\}}\mathbb{1}_{\{\hat{\bs{y}}_{\bs{\theta}}(\bs{x}_i)\in R_j(\bs{\tau})\}}},\; \mathrm{FN}_j(\bs{\tau},\bs{\theta}) = \sum_{i=1}^n{\mathbb{1}_{\{\bs{y}_i= \bs{e}_j\}}\mathbb{1}_{\{\hat{\bs{y}}_{\bs{\theta}}(\bs{x}_i)\notin R_j(\bs{\tau})\}}}.
        \end{split}
\end{equation*}
Note that $\hat{\bs{y}}$ is assigned to the positive class for a unique $\mathrm{CM}_j$ only, being negative in the other confusion matrices corresponding to different indices.

Then, the contribution of all confusion matrices can be combined in a vector
\begin{equation}\label{eq:vec_combined}
    \bs{s}(\bs{\tau},\bs{\theta})=( s(\mathrm{CM}_1(\bs{\tau},\bs{\theta})),\dots,s(\mathrm{CM}_m(\bs{\tau},\bs{\theta})) ),
\end{equation}
and summarized by taking, e.g., the mean value
\begin{equation*}
    s_{\textrm{mean}}(\bs{\tau},\bs{\theta})=\frac{1}{m}\lVert \bs{s}(\bs{\tau},\bs{\theta})\lVert_1 .
\end{equation*}
Averaging over the contributions of all binary scores, as considered here, is the so-called \textit{macro} setting, which is particularly useful in presence of unbalancing in classes' distribution.
Then, a posteriori maximization can be carried out after the training phase by computing (cf. \eqref{eq:a_posteriori})
\begin{equation}\label{eq:a_posteriori_multi}
    \max_{\bs{\tau}\in S_m}s_{\textrm{mean}}(\bs{\tau},\bs{\theta}^\star).
\end{equation}
We sum up our score maximization scheme in Algorithm \ref{alg:tuning}, and in \ref{sec:results} we present several classification tests that show the effectiveness of the algorithm in improving the performance of trained networks.
\begin{algorithm}[h!]
	\caption{Multidimensional threshold tuning}\label{alg:tuning}
	\begin{algorithmic}[1]
		\State \textbf{Input:} True labels $\{\bs{y}_1,\dots,\bs{y}_n\}$, corresponding predictions from trained network $\{\hat{\bs{y}}_{\bs{\theta}^\star}(\bs{x}_1),\dots,\hat{\bs{y}}_{\bs{\theta}^\star}(\bs{x}_n)\}$ (training or validation set), classification score $s$
		\State \textbf{Output:} Best threshold $\bs{\tau}^\star$
		\State Sample $\bs{\tau}_1,\dots,\bs{\tau}_M$ threshold values on the simplex $S_m$ (e.g.\ uniform grid)
		\For{$k=1,\dots,M$}
        \State Construct $\mathrm{CM}_j(\bs{\tau}_k,\bs{\theta}^\star)$ for each $j=1,\dots,m$
        \State Evaluate $s(\mathrm{CM}_j(\bs{\tau}_k,\bs{\theta}^\star))$ for each $j=1,\dots,m$
        \State Calculate $s_{\textrm{mean}}(\bs{\tau}_k,\bs{\theta}^\star)=\frac{1}{m}\sum_{j=1}^m s(\mathrm{CM}_j(\bs{\tau}_k,\bs{\theta}^\star))$
		\EndFor
		\State Compute $\bs{\tau}^\star=\bs{\tau}_{k^\star}$ where $k^\star= \argmax_{k=1,\dots,M}s_{\textrm{mean}}(\bs{\tau}_k,\bs{\theta}^\star)$
	\end{algorithmic}
\end{algorithm}

\section{ROC analysis in the simplex}\label{sec:roc}

When dealing with multiple classes in the ROC analysis, the classical approach consists of building many ROC curves, one for each class considered positive \textit{versus} the other classes considered negative, which may be then aggregated a posteriori to summarize the overall performance of the classifier. Formally, this approach can be formalized as follows. For each class $i=1,\dots,m$, the multiclass output is binarized as
\begin{equation*}
   \hat{\bs{y}}_{\bs{\theta}} = (\hat{y}_{\bs{\theta}}^1,\dots,\hat{y}_{\bs{\theta}}^m) \Longrightarrow  \hat{\bs{y}}_{\bs{\theta},i}=\bigg(\hat{y}_{\bs{\theta}}^i,\sum_{j\neq i}\hat{y}_{\bs{\theta}}^j\bigg),
\end{equation*}
so that the native binary procedure can be applied separately for each $\hat{\bs{y}}_{\bs{\theta},i}$, $i=1,\dots,m$.

\begin{figure}[htbp]
	\centering
	\includegraphics[width=\linewidth]{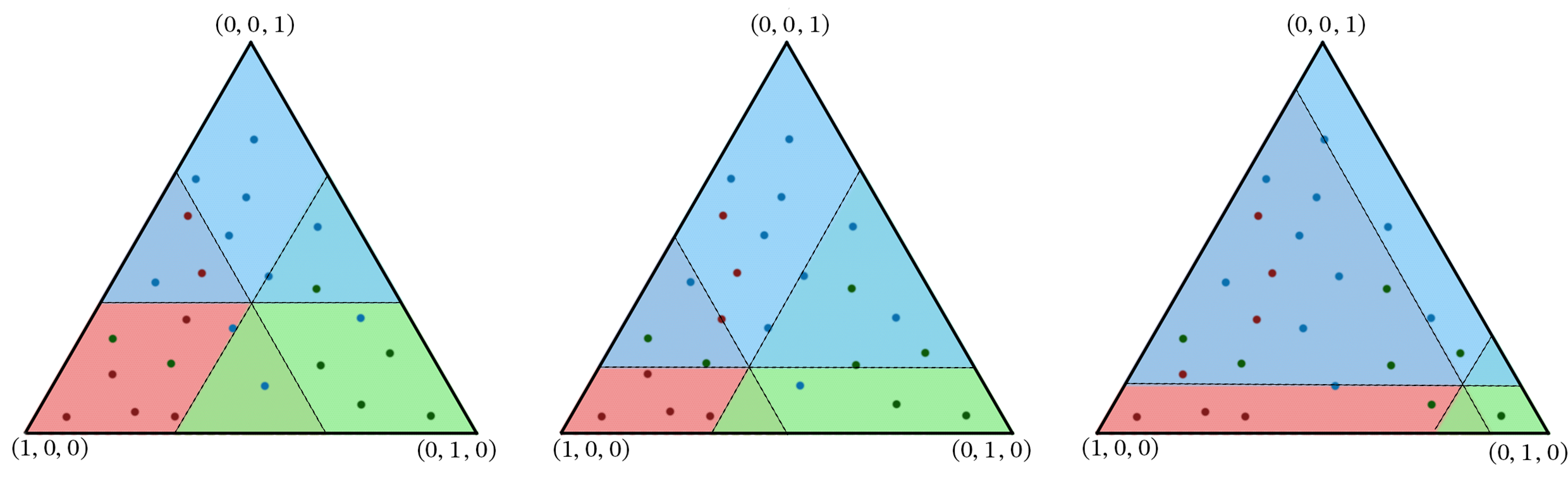} 
	\caption{For $m=3$, the three regions $\bar{R}_1(\bs{\tau})$ (red), $\bar{R}_2(\bs{\tau})$ (green) and $\bar{R}_3(\bs{\tau})$ (blue). From left to right: $\bs{\tau}=(1/3,1/3,1/3)$, $\bs{\tau}=(1/2,1/3,1/6)$, $\bs{\tau}=(1/8,3/4,1/8)$ (big black dot). Differently with respect to \ref{fig:simplices}, here each region is unrelated to the others and overlapping is allowed.}
	\label{fig:simplices_overlapped}
\end{figure}

To better understand the discrepancy that this OvR approach determines with respect to the true multiclass output offered by the classifier, we can take advantage of our simplex framework. In fact, in $S_m$, this is equivalent to considering the classification collection produced by the following classification regions:
\begin{equation*}
    \bar{R}_j(\bs{\tau})=\{\bs{z}\in S_m\:|\: z^j>\tau^j\},\; j=1,\dots,m.
\end{equation*}
Note that the collection $\bar{R}_1(\bs{\tau}),\dots,\bar{R}_m(\bs{\tau})$ satisfies in fact the requirements of Definition \ref{def:region}, but it is not proper according to Definition \ref{def:region_proper}, as $\bar{R}_i(\bs{\tau})\cap \bar{R}_j(\bs{\tau})\neq\emptyset$ for $i,j=1,\dots,m$ (see \ref{fig:simplices_overlapped} and cf.\ \ref{fig:simplices}).
Therefore, the major flaws of this approach are:
\begin{itemize}
    \item 
    Each classification region does not relate to the others, completely ignoring the information provided by the classifier that \textit{jointly} distinguish among multiple classes, leading to its incomplete evaluation.
    \item 
    Since the collection is not proper, we are formally considering a hybrid \\multiclass-multilabel framework, where an element might belong to more than one class, which substantially differs with respect to the classification process carried out in the training phase.
\end{itemize}
Thanks to our threshold-based framework, fixed a score $s$ and with the choice \eqref{eq:natural}, we recall that we can construct $m$ classification matrices for a fixed threshold value $ \boldsymbol{\tau}\in S_m$, and derive a vector $\boldsymbol{s}$ of corresponding score values (see \eqref{eq:vec_combined}). Therefore, focusing on FPR and TPR scores, we can obtain a map
\begin{equation}\label{eq:mappa}
    \boldsymbol{\tau}\longrightarrow \{\boldsymbol{fpr}( \boldsymbol{\tau}),\boldsymbol{tpr}( \boldsymbol{\tau})\},
\end{equation}
where we omitted the dependence on the weights of the network.

To clarify this procedure with an example, let us again consider the setting in \ref{fig:simplices}.
\begin{itemize}
    \item 
    For $\bs{\tau}=(1/3,1/3,1/3)$, we get $\boldsymbol{fpr}(\boldsymbol{\tau})=(4/17,2/17,1/14)$ and $\boldsymbol{tpr}(\boldsymbol{\tau})=(6/7,5/7,6/10)$.
    \item 
    For $\bs{\tau}=(1/2,1/3,1/6)$, we have $\boldsymbol{fpr}(\boldsymbol{\tau})=(2/17,2/17,4/14)$ and $\boldsymbol{tpr}(\boldsymbol{\tau})=(4/7,4/7,8/10)$.
    \item 
    For $\bs{\tau}=(1/8,3/4,1/8)$, we get $\boldsymbol{fpr}(\boldsymbol{\tau})=(7/17,0,3/14)$ and $\boldsymbol{tpr}(\boldsymbol{\tau})=(6/7,1/7,7/10)$.
\end{itemize}
By computing the scores for many threshold values $\bs{\tau}_1,\dots,\bs{\tau}_M$, we can obtain multiple sets
\begin{equation}\label{eq:couples}
    \{\boldsymbol{fpr}( \boldsymbol{\tau}_1),\boldsymbol{tpr}( \boldsymbol{\tau}_1)\},\dots,\{\boldsymbol{fpr}( \boldsymbol{\tau}_M),\boldsymbol{tpr}( \boldsymbol{\tau}_M)\}.
\end{equation}
Now, we need to process \eqref{eq:couples} to extract information on the performance of the classifier. Fixed $\boldsymbol{\tau}_k$, each couple $\boldsymbol{fpr}( \boldsymbol{\tau}_k),\boldsymbol{tpr}( \boldsymbol{\tau}_k)$ might be framed in $[0,1]^m\times[0,1]^m$ space. Consequently, by varying $\boldsymbol{\tau}$ across the simplex and in light of \eqref{eq:mappa}, one could interpret \eqref{eq:couples} as a set of achievable points that form a surface in the reference space. However, this surface can not be directly visualized for $m > 3$, and interpreting its properties for model evaluation purposes is an intricate task. Alternatively, we reduce our setting to 2D \textit{projections}, that is, we collect the $i$-th coordinates of each vector to determine 2D points as
\begin{equation*}
    (fpr^j( \boldsymbol{\tau}_1),tpr^j( \boldsymbol{\tau}_1)),\dots, (fpr^j( \boldsymbol{\tau}_M),tpr^j( \boldsymbol{\tau}_M)),
\end{equation*}
and thus define for each $j=1,\dots,m$ a ROC cloud in the usual FPR/TPR plane. Then, the information can be summarized by considering the following Distance From Point (DFP) measure:
\begin{equation}\label{eq:DFP}
    \textrm{DFP}_j=\frac{1}{M}\sum_{k=1}^M \left\lVert  (fpr^j( \boldsymbol{\tau}_k),tpr^j( \boldsymbol{\tau}_k))-(0,1)\right\rVert_1,\quad j=1,\dots,m,
\end{equation}
being $(0,1)$ the FPR/TPR output of a perfect classification, and the metrics of all classes can be combined in a unique DFP measure by taking, e.g., the mean value; see Algorithm \ref{alg:roc} for a formalization of the proposed procedure.

\begin{algorithm}[h!]
	\caption{ROC analysis in the multidimensional simplex}\label{alg:roc}
	\begin{algorithmic}[1]
		\State \textbf{Input:} True labels $\{\bs{y}_1,\dots,\bs{y}_n\}$, corresponding predictions from trained network $\{\hat{\bs{y}}_{\bs{\theta}^\star}(\bs{x}_1),\dots,\hat{\bs{y}}_{\bs{\theta}^\star}(\bs{x}_n)\}$ (training or validation set)
		\State \textbf{Output:} ROC clouds, ROC-DFP measure
		\State Sample $\bs{\tau}_1,\dots,\bs{\tau}_M$ threshold values on the simplex $S_m$ (e.g.\ uniform grid)
		\For{$k=1,\dots,M$}
        \State Calculate the vectors $\boldsymbol{fpr}( \boldsymbol{\tau}_k),\boldsymbol{tpr}( \boldsymbol{\tau}_k)$
		\EndFor
        \For{$j=1,\dots,m$}
		\State Set the ROC cloud $(fpr^j( \boldsymbol{\tau}_1),tpr^j( \boldsymbol{\tau}_1)),\dots, (fpr^j( \boldsymbol{\tau}_M),tpr^j( \boldsymbol{\tau}_M))$
        \State Compute the measure $\textrm{DFP}_j$ as in \eqref{eq:DFP}
        \EndFor
        \State Calculate the summarizing $\textrm{DFP}=\frac{1}{m}\sum_{j=1}^m \textrm{DFP}_j$
	\end{algorithmic}
\end{algorithm}

Note that each $\textrm{DFP}_j$ consists of a discrete $L_1$-Wasserstein distance, that is, the transport measure that computes the mean value of the $L_1$ distances covered to \textit{move} the FPR/TPR couples onto the $(0,1)$ point. The choice for the $L_1$ distance, also called the Manhattan distance, is motivated by the fact that the FPR/TPR couples lying in the \textit{random bisector} of the ROC plane are then equidistant from $(0,1)$ and at maximum distance. Consequently, while $\textrm{DFP}_j=0$ is the perfect result, $\textrm{DFP}_j=1$ is the maximum value in a random setting.

We remark that the crucial difference of our approach from the classical one is that here the FPR/TPR couples that lean the ROC analysis are not the result of separate threshold evaluations but are intertwined among the classes, as they rely on the same multidimensional threshold value, being natively multiclass and more coherent with the actual output of the multiclass model; see \ref{tab:roc_vs_simplexroc}.
\begin{table}[h]
\centering
\resizebox{\textwidth}{!}{%
\begin{tabular}{|r|c|c|}
\hline
 & \textbf{Classical} & \textbf{Our approach} \\
\hline
\textbf{Threshold(s)} & \makecell{Separate independent thresholds\\ for each class} 
             & \makecell{Unique multidimensional threshold\\ for all classes}  \\
\hline
\textbf{Outputs}      & \makecell{Unrelated FPR/TPR\\ OvR couples}  
             & \makecell{Intertwined FPR/TPR\\ OvR couples}\\
\hline
\textbf{Analysis tool} & OvR ROC curves  & OvR ROC clouds  \\
\hline
\textbf{Summary metric} & AUC  & DFP  \\
\hline
\end{tabular}%
}
\caption{Key differences between the classical multidimensional ROC analysis and our proposed approach in the threshold-based framework in the simplex.}
\label{tab:roc_vs_simplexroc}
\end{table}

\section{Results}\label{sec:results}

The experiments for this project can be run on consumer-grade hardware. In our case, they were conducted on a desktop PC equipped with 64 GB of RAM, a 16-core AMD Ryzen 5950X CPU, and an NVIDIA RTX 3070 GPU with 8 GB of VRAM.

The threshold tuning python package is available at
\begin{equation*}
    \texttt{https://github.com/edoardolegnaro/SimplexTools}.
\end{equation*}

\subsection{Multiclass threshold optimization}\label{sec:test_soglia}

\begin{figure}
	\centering
\includegraphics[width=\textwidth]{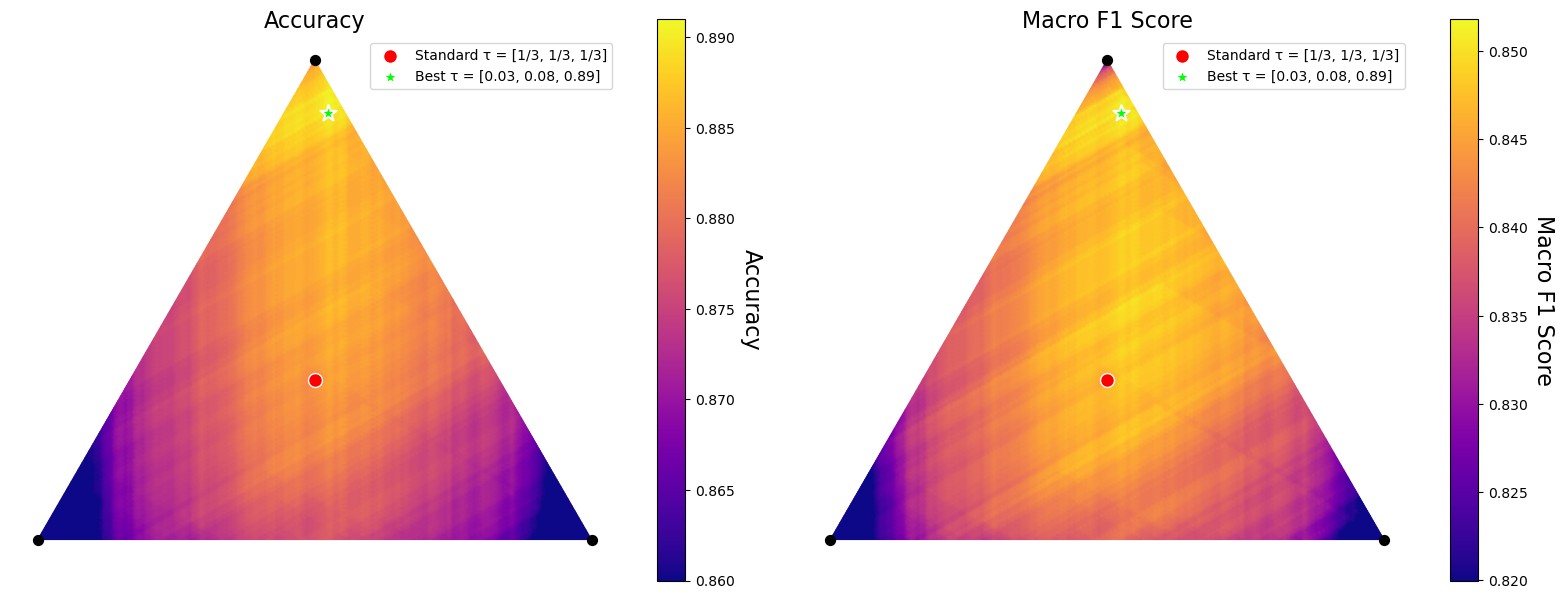}
	\caption{Accuracy (left) and Macro F1 Score (right) heatmaps across the simplex for the validation set. The color intensity represents the metric's value, with lighter shades indicating higher performance. The standard threshold $\bs{\tau} = (1/3, 1/3, 1/3)$ is marked with a red circle, while the threshold giving the best score is indicated by a green star, corresponding to $\bs{\tau}^\star = (0.00, 0.06, 0.94)$ for accuracy and $\bs{\tau}^\star = (0.01, 0.08, 0.91)$ for the Macro F1 Score. Here, the grid of evaluation thresholds consists of $M=20301$ samples.}
	\label{fig:tang_score}
\end{figure}

In this subsection, our purpose is to show the effectiveness of the proposed threshold tuning approach (see Algorithm \ref{alg:tuning}) in refining the performance of a classifier trained with the weighted categorical CE. Different datasets with varying numbers of labels $m$ are considered. The distributions of the classes over training, validation and test sets for the datasets are provided in \ref{tab:classes_counts}.
\begin{table}
\centering
\resizebox{\textwidth}{!}{
\begin{tabular}{crccc}
\toprule
\textbf{Dataset} & \textbf{Class} & \textbf{Train} & \textbf{Validation} & \textbf{Test} \\
\midrule
\multirow{4}{*}{SOLAR-STORM1} & $\alpha$ & 3821 (33.01\%) & 888 (30.68\%) & 567 (48.38\%) \\
& $\beta$ & 5691 (49.17\%) & 1662 (57.43\%) & 496 (42.32\%) \\
& $\beta X$ & 2063 (17.82\%) & 344 (11.89\%) & 109 (9.30\%) \\
\cmidrule(lr){2-5}
& \textbf{Total} & \textbf{11575} & \textbf{2894} & \textbf{1172} \\
\midrule
\multirow{5}{*}{OCTMNIST} & choroidal neovascularization & 33484 (34.35\%) & 3721 (34.35\%) & 250 (25.00\%) \\
& diabetic macular edema & 10213 (10.48\%) & 1135 (10.48\%) & 250 (25.00\%) \\
& drusen & 7754 (7.95\%) & 862 (7.96\%) & 250 (25.00\%) \\
& normal & 46026 (47.22\%) & 5114 (47.21\%) & 250 (25.00\%) \\
\cmidrule(lr){2-5}
& \textbf{Total} & \textbf{97477} & \textbf{10832} & \textbf{1000} \\
\midrule
\multirow{10}{*}{PATHMNIST} & adipose & 9366 (10.41\%) & 1041 (10.41\%) & 1338 (18.64\%) \\
& background & 9509 (10.57\%) & 1057 (10.57\%) & 847 (11.80\%) \\
& debris & 10360 (11.51\%) & 1152 (11.52\%) & 339 (4.72\%) \\
& lymphocytes & 10401 (11.56\%) & 1156 (11.56\%) & 634 (8.83\%) \\
& mucus & 8006 (8.90\%) & 890 (8.90\%) & 1035 (14.42\%) \\
& smooth muscle & 12182 (13.54\%) & 1354 (13.53\%) & 592 (8.25\%) \\
& normal colon mucosa & 7886 (8.76\%) & 877 (8.77\%) & 741 (10.32\%) \\
& cancer-associated stroma & 9401 (10.45\%) & 1045 (10.45\%) & 421 (5.86\%) \\
& colorectal adenocarcinoma epithelium & 12885 (14.32\%) & 1432 (14.31\%) & 1233 (17.17\%) \\
\cmidrule(lr){2-5}
& \textbf{Total} & \textbf{89996} & \textbf{10004} & \textbf{7180} \\
\midrule
\multirow{11}{*}{FASHIONMNIST} & T-shirt/top & 4836 (10.08\%) & 1210 (10.08\%) & 1000 (10.00\%) \\
& Trouser & 4812 (10.03\%) & 1264 (10.53\%) & 1000 (10.00\%) \\
& Pullover & 4773 (9.94\%) & 1207 (10.06\%) & 1000 (10.00\%) \\
& Dress & 4774 (9.95\%) & 1146 (9.55\%) & 1000 (10.00\%) \\
& Coat & 4800 (10.00\%) & 1243 (10.36\%) & 1000 (10.00\%) \\
& Sandal & 4833 (10.07\%) & 1180 (9.83\%) & 1000 (10.00\%) \\
& Shirt & 4867 (10.14\%) & 1183 (9.86\%) & 1000 (10.00\%) \\
& Sneaker & 4786 (9.97\%) & 1176 (9.80\%) & 1000 (10.00\%) \\
& Bag & 4749 (9.89\%) & 1194 (9.95\%) & 1000 (10.00\%) \\
& Ankle boot & 4770 (9.94\%) & 1197 (9.98\%) & 1000 (10.00\%) \\
\cmidrule(lr){2-5}
& \textbf{Total} & \textbf{48000} & \textbf{12000} & \textbf{10000} \\
\bottomrule
\end{tabular}%
}
\caption{Classes distributions for SOLAR-STORM1, OCTMNIST, PATHMNIST, and FASHIONMNIST datasets.}
\label{tab:classes_counts}
\end{table}
First, we consider the task of classifying solar active regions (ARs) into their magnetic type \cite{guastavino2022implementation}.
We take the SOLAR-STORM1 dataset provided in \cite{fang2019deep}, which consists of images of sunspots from the Helioseismic and Magnetic Imager (HMI) instrument on Solar Dynamics Observatory (SDO) satellite \cite{pesnell2012solar,scherrer2012helioseismic}. The classes are three: $\alpha$ (unipolar sunspots), $\beta$ (bipolar sunspots) and $\beta X$ (complex sunspots), with unbalanced classes distributions.
We consider a Data Efficient Image Transformer (DeiT, \cite{touvron2021training})
with about $86.6$M parameters (the \texttt{deit\_base\_patch16\_224} implementation
from \cite{rw2019timm}) trained following \cite{legnaro2025deep}.
The scores found are reported in \ref{tab:performance_metrics_thresh} (top), and in \ref{fig:tang_score} we show the distribution of the scores obtained by varying the threshold in the $2$-simplex.

The remaining experiments in \ref{tab:performance_metrics_thresh} are carried out by training a \texttt{resnet18} model, also making use of standard tools such as: data augmentations, mixed precision training, Adam optimizer and an early stopping rule. The network weights are restored to the best validation case in order to make predictions for the test. We show the results obtained for OCTMNIST \cite{octmnist}, PATHMNIST \cite{pathmnist} \cite{medmnistv1,medmnistv2} and FASHIONMNIST \cite{xiao2017fashion}.

\begin{table}[htbp]
\centering
\scriptsize
\resizebox{\textwidth}{!}{%
\begin{tabular}{@{}ccccccc@{}}
\toprule
\textbf{Dataset} & \textbf{Metric} & \textbf{Set} & \textbf{Method} & \textbf{Value} & \textbf{Delta} & \textbf{Validation $\bs{\tau}^\star$} \\
\midrule

\multirow{8}{*}{SOLAR-STORM1 ($m=3$)} 
& \multirow{4}{*}{Accuracy} 
    & \multirow{2}{*}{Validation} & Argmax & 0.8829 & \multirow{2}{*}{0.0069} & \multirow{4}{*}{\centering $(0.00 \quad 0.06 \quad 0.94)$} \\ 
&   &                             & Tuned  & 0.8898 &                         &                                                       \\
\cmidrule(lr){3-6} 
&   & \multirow{2}{*}{Test}       & Argmax & 0.8831 & \multirow{2}{*}{0.0154} &                                                       \\
&   &                             & Tuned  & 0.8985 &                         &                                                       \\
\cmidrule(lr){2-7} 
& \multirow{4}{*}{F1 Score} 
    & \multirow{2}{*}{Validation} & Argmax & 0.8462 & \multirow{2}{*}{0.0041} & \multirow{4}{*}{\centering $(0.01 \quad 0.08 \quad 0.91)$} \\ 
&   &                             & Tuned    & 0.8503 &                         &                                                       \\
\cmidrule(lr){3-6}
&   & \multirow{2}{*}{Test}       & Argmax & 0.8527 & \multirow{2}{*}{0.0252} &                                                       \\
&   &                             & Tuned    & 0.8779 &                         &                                                       \\
\midrule 

\multirow{8}{*}{OCTMNIST ($m=4$)} 
& \multirow{4}{*}{Accuracy} 
    & \multirow{2}{*}{Validation} & Argmax & 0.8816 & \multirow{2}{*}{0.0297} & \multirow{4}{*}{\centering $(0.00 \quad 0.24 \quad 0.64 \quad 0.12)$} \\
&   &                             & Tuned  & 0.9113 &                         &                                                               \\
\cmidrule(lr){3-6}
&   & \multirow{2}{*}{Test}       & Argmax & 0.8420 & \multirow{2}{*}{0.0000} &                                                               \\
&   &                             & Tuned  & 0.8420 &                         &                                                               \\

\cmidrule(lr){2-7}
& \multirow{4}{*}{F1 Score} 
    & \multirow{2}{*}{Validation} & Argmax & 0.8214 & \multirow{2}{*}{0.0179} & \multirow{4}{*}{\centering $(0.16 \quad 0.24 \quad 0.46 \quad 0.14)$} \\
&   &                             & Tuned  & 0.8393 &                         &                                                               \\
\cmidrule(lr){3-6}
&   & \multirow{2}{*}{Test}       & Argmax & 0.8384 & \multirow{2}{*}{-0.0566} &                                                               \\
&   &                             & Tuned  & 0.7818 &                         &                                                               \\
\midrule

\multirow{8}{*}{PATHMNIST ($m=9$)} 
& \multirow{4}{*}{Accuracy} 
    & \multirow{2}{*}{Validation} & Argmax & 0.9705 & \multirow{2}{*}{0.0006} & \multirow{4}{*}{\centering \begin{tabular}{@{}c@{}}$(0.00 \quad 0.56 \quad 0.44 \quad 0.00 \quad 0.00$ \\ $0.00 \quad 0.00 \quad 0.00 \quad 0.00)$\end{tabular}} \\
&   &                             & Tuned  & 0.9711 &                         &                                                               \\
\cmidrule(lr){3-6}
&   & \multirow{2}{*}{Test}       & Argmax & 0.9125 & \multirow{2}{*}{0.0000} &                                                               \\
&   &                             & Tuned  & 0.9125 &                         &                                                               \\
\cmidrule(lr){2-7}
& \multirow{4}{*}{F1 Score} 
    & \multirow{2}{*}{Validation} & Argmax & 0.9705 & \multirow{2}{*}{0.0007} & \multirow{4}{*}{\centering \begin{tabular}{@{}c@{}}$(0.00 \quad 0.56 \quad 0.44 \quad 0.00 \quad 0.00$ \\ $0.00 \quad 0.00 \quad 0.00 \quad 0.00)$\end{tabular}} \\
&   &                             & Tuned    & 0.9712 &                         &                                                               \\
\cmidrule(lr){3-6}
&   & \multirow{2}{*}{Test}       & Argmax & 0.8760 & \multirow{2}{*}{0.0016} &                                                               \\
&   &                             & Tuned    & 0.8776 &                         &                                                               \\
\midrule

\multirow{8}{*}{FASHIONMNIST ($m=10$)} 
& \multirow{4}{*}{Accuracy} 
    & \multirow{2}{*}{Validation} & Argmax & 0.9838 & \multirow{2}{*}{0.0004} & \multirow{4}{*}{\centering \begin{tabular}{@{}c@{}}$(0.00 \quad 0.67 \quad 0.00 \quad 0.00 \quad 0.33$ \\ $0.00 \quad 0.00 \quad 0.00 \quad 0.00 \quad 0.00)$\end{tabular}} \\
&   &                             & Tuned  & 0.9842 &                         &                                                               \\
\cmidrule(lr){3-6}
&   & \multirow{2}{*}{Test}       & Argmax & 0.9347 & \multirow{2}{*}{0.0000} &                                                               \\
&   &                             & Tuned  & 0.9347 &                         &                                                               \\
\cmidrule(lr){2-7}
& \multirow{4}{*}{F1 Score} 
    & \multirow{2}{*}{Validation} & Argmax & 0.9838 & \multirow{2}{*}{0.0004} & \multirow{4}{*}{\centering \begin{tabular}{@{}c@{}}$(0.00 \quad 0.67 \quad 0.00 \quad 0.00 \quad 0.33$ \\ $0.00 \quad 0.00 \quad 0.00 \quad 0.00 \quad 0.00)$\end{tabular}} \\
&   &                             & Tuned    & 0.9842 &                         &                                                               \\
\cmidrule(lr){3-6}
&   & \multirow{2}{*}{Test}       & Argmax & 0.9346 & \multirow{2}{*}{0.0000} &                                                               \\
&   &                             & Tuned    & 0.9346 &                         &                                                               \\
\bottomrule
\end{tabular}%
}
\caption{Performance metrics for different datasets with corresponding best validation threshold $\bs{\tau}^\star$.}
\label{tab:performance_metrics_thresh}
\end{table}

To provide a more comprehensive evaluation, besides the standard \textit{argmax} rule, the performance of the proposed multidimensional threshold tuning (``Tuned'') was benchmarked against  two other recently developed decision rules from the literature: the Fréchet mean \cite{roberts2023geometry} and the $\varepsilon$-inflated \textit{argmax} \cite{soloff2024building}.
A notable characteristic of the $\varepsilon$-inflated \textit{argmax} is its capacity to predict a set of labels rather than a single classification output. Consequently, its performance was assessed using two specific metrics:
\begin{itemize}
    \item \textbf{Coverage}, which considers a prediction correct if the true label is contained within the predicted set.
    \item \textbf{Singleton}, which calculates accuracy in the conventional manner but is restricted to predictions where the output set consists of a single label.
\end{itemize}

The experimental results are summarised in \ref{tab:decision_rules_comparison}. As indicated by the ``Average Set Size'' metric, non-singleton predictions were infrequent in the experiments.

\begin{table}[H]
\centering
\resizebox{\textwidth}{!}{%
\begin{tabular}{@{}cccccccc@{}}
\toprule
\textbf{Dataset} & \textbf{Set} & \textbf{Argmax} & \textbf{Tuned (our)} & \textbf{Fréchet} & \multicolumn{3}{c}{\textbf{$\varepsilon$-inflated Argmax}} \\
\cmidrule(l){6-8}
& & & & & \textbf{Coverage} & \textbf{Singleton} & \textbf{Avg Set Size} \\
\midrule
\multirow{2}{*}{SOLAR-STORM1} & Validation & 0.8829 & 0.8898 & 0.8829 & 0.8856 & 0.8818 & 1.0038 \\
& Test & 0.8831 & 0.8985 & 0.8831 & 0.8882 & 0.8805 & 1.0077 \\
\midrule
\multirow{2}{*}{OCTMNIST} & Validation & 0.8816 & 0.9093 & 0.8816 & 0.8883 & 0.8744 & 1.0147 \\
& Test & 0.8420 & 0.8420 & 0.8420 & 0.8510 & 0.8340 & 1.0170 \\
\midrule
\multirow{2}{*}{PATHMNIST} & Validation & 0.9705 & 0.9711 & 0.9705 & 0.9709 & 0.9695 & 1.0020 \\
& Test & 0.9125 & 0.9125 & 0.9125 & 0.9149 & 0.9102 & 1.0054 \\
\midrule
\multirow{2}{*}{FASHIONMNIST} & Validation & 0.9838 & 0.9842 & 0.9809 & 0.9811 & 0.9804 & 1.0008 \\
& Test & 0.9347 & 0.9347 & 0.9347 & 0.9361 & 0.9339 & 1.0027 \\
\bottomrule
\end{tabular}%
}
\caption{Comparison of accuracy for different decision rules. For the $\varepsilon$-inflated argmax, ``Coverage'' indicates correctness if the true label is in the predicted set, while ``Singleton'' is the accuracy on single-label predictions.}
\label{tab:decision_rules_comparison}
\end{table}

\subsection{ROC analysis}\label{sec:test_roc}
To demonstrate the proposed ROC analysis framework, we consider two datasets: SOLAR-STORM1 ($m{=}3$) and OCTMNIST ($m{=}4$). 
On each test set, we vary the multidimensional threshold $\boldsymbol{\tau}$ uniformly over the simplex $S_m$ computing the corresponding classwise $(\mathrm{FPR},\mathrm{TPR})$ pairs. 
The resulting point sets form the \emph{ROC clouds}, which represent all achievable operating points under a single, coherent multiclass decision rule. 
Unlike standard One-vs-Rest (OvR) curves, which vary independent thresholds per class, the ROC clouds reflect realistic joint trade-offs among classes.
For reference, the orange curves in the figures show the traditional OvR ROC curves. These are computed as follows: for each class $j$, scores $\hat y^j$ are compared against a binarized label ($j$ vs.\ rest) and the threshold on $\hat y^j$ is swept to trace $(\mathrm{FPR},\mathrm{TPR})$.

Typically, the ROC clouds lie below or on par with the OvR curves, as expected since the latter represent the best possible performance when varying thresholds independently per class.
Nevertheless, as seen in the SOLAR-STORM1 case especially for class beta, points in the cloud can exceed the OvR curve, showing that the simplex-based rule provides better performance compared to argmax, as confirmed also by the result in \ref{tab:performance_metrics_thresh}.
Instead, in the OCTMNIST case, the same table shows that the simplex decision rule does not give an edge in the test scores, and consequently the ROC clouds are mostly below or on par with the OvR curves.

The DFP computations for these datasets are summarized in \ref{tab:dfp_values} and quantify the visual gaps in the clouds: in SOLAR-STORM1, \emph{beta} (0.2297) is closest to $(0,1)$ while \emph{alpha} (0.3312) is farthest; in OCTMNIST, \emph{Drusen} (0.2264) and \emph{Normal} (0.2262) are most separable, whereas \emph{CNV} (0.3042) is most challenging.

\begin{table}[h]

\centering
\small{
\begin{tabular}{rccccc}
\toprule
\textbf{Dataset} & \textbf{Class 1} & \textbf{Class 2} & \textbf{Class 3} & \textbf{Class 4} & \textbf{Overall DFP} \\
\midrule
SOLAR-STORM1 & 0.3312 & 0.2297 & 0.2471 & -- & \textbf{0.2693} \\
\midrule
OCTMNIST & 0.3042 & 0.2735 & 0.2264 & 0.2262 & \textbf{0.2576} \\
\bottomrule
\end{tabular}
}
\caption{Per-class and overall DFP values for SOLAR-STORM1 and OCTMNIST. 
Smaller DFP values correspond to ROC clouds closer to the ideal top-left corner $(0,1)$, indicating better class separability.}
\label{tab:dfp_values}
\end{table}

\begin{figure}[h]
\centering
\includegraphics[width=\textwidth]{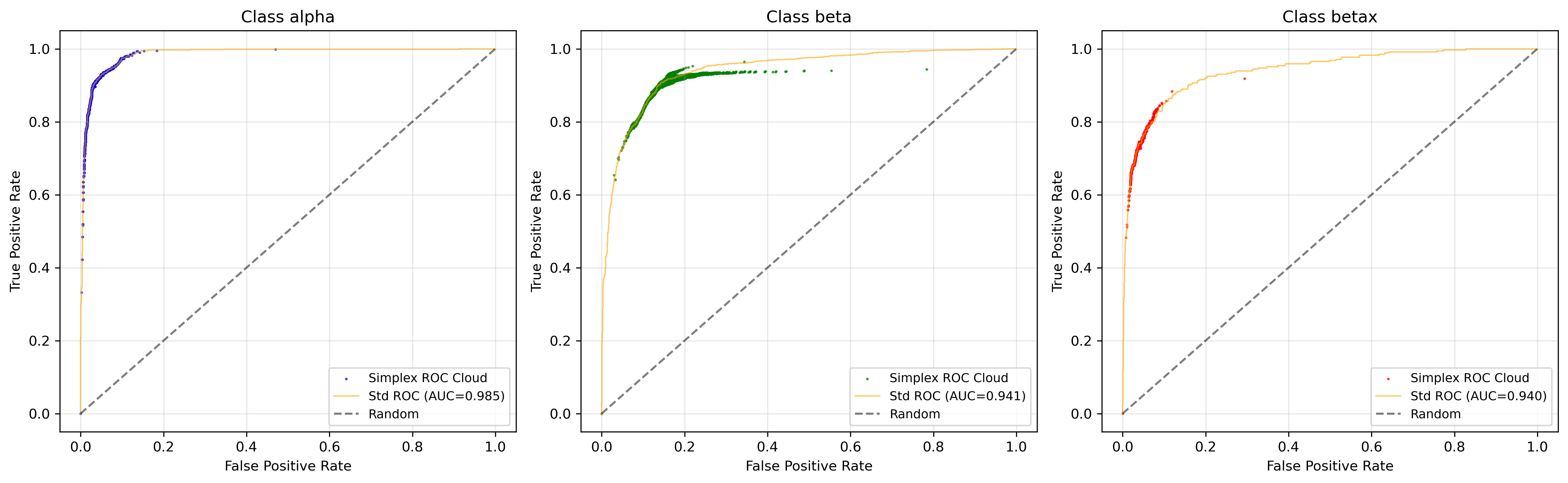}

\vspace{0.6em}

\includegraphics[width=\textwidth]{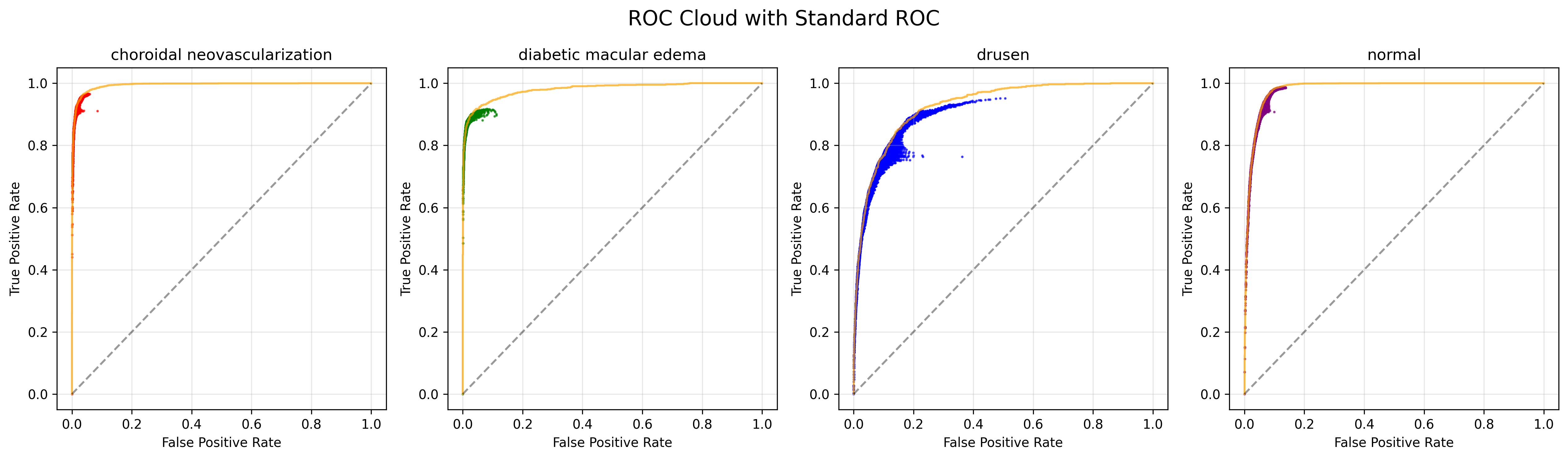}
\caption{ROC clouds with OvR ROC curves (orange). Top: SOLAR-STORM1. Bottom: OCTMNIST (validation set).}
\label{fig:roc_clouds_combined}
\end{figure}

\section{Discussion}\label{sec:discussion}

The experiments carried out in \ref{sec:test_soglia} show the effectiveness of the proposed threshold tuning approach in enhancing and improving the performance of classification networks. 
As it can be applied a posteriori to any neural classifier, this procedure has the potential to become a common practice, just as classical one-dimensional threshold tuning in a binary setting. 
As intuition suggests from the binary case, such multidimensional threshold tuning could be especially meaningful in unbalanced settings. Note that in \ref{fig:tang_score} the best threshold turns out to be \textit{close} to the minority class, as one would expect in a binary task.

We remark that the effectiveness of tuning the threshold on the validation set to improve the performance on the test set is strongly influenced by the classes' distribution. 
Indeed, even if, according to machine learning theory, validation and test sets should be characterized by an analogous classes' distribution \cite{Vapnik1998}, this may not take place in fact, thus determining a calibration discrepancy between validation and test sets. In our tests reported in \ref{tab:performance_metrics_thresh}, we actually obtained the worst result, i.e., a worse performance on the test set with respect to the argmax rule, in the case where the classes' distribution on the test set deviates the most from the validation one, that is the case of the OCTMNIST dataset. Besides reasonable oscillations in performance due to random factors, initializations, and train/test splits, differences between validation and test compositions are usually responsible for unsatisfactory calibration results.

Furthermore, we compared the performance of our proposed tuning strategies with respect to other state-of-the-art decision rules. Results in \ref{tab:decision_rules_comparison} demonstrate the effectiveness of the Tuned threshold approach, which consistently matches or exceeds the performance of both the baseline \textit{argmax} and the alternative decision rules.

A principal advantage of the Tuned strategy is that, by design, it can only improve upon or equate the performance of the classical \textit{argmax} rule on the validation set. This is because the \textit{argmax} condition corresponds to the barycenter of the simplex, which is one of the candidate thresholds evaluated during the tuning process. In contrast, other competitive strategies, such as the Fréchet mean or the $\varepsilon$-inflated \textit{argmax}, may yield occasional performance degradation compared to the standard procedure. As is standard, the generalization of performance to the test set depends on factors such as the congruence of class distributions between the validation and test splits. Regarding the $\varepsilon$-inflated \textit{argmax}, while its ``Coverage'' score is sometimes marginally higher than that of Tuned and other methods, this is attributable to its set-based prediction mechanism. When restricted to single-label outputs (``Singleton'' accuracy), the $\varepsilon$-inflated \textit{argmax} generally underperforms compared to Tuned. These findings underscore the robustness and practical significance of the proposed Tuned threshold framework.

As far as the proposed ROC analysis is concerned, 
\ref{fig:roc_clouds_combined} visualizes the proposed ROC clouds alongside the standard OvR curves.
As expected, clouds typically lie below or on par with OvR, because OvR sweeps an unconstrained per-class threshold and thus traces an idealized, class-isolated frontier for the considered class while ignoring the consequences for the others. In contrast, each point in a ROC cloud is produced by a \emph{single} multiclass threshold $\boldsymbol{\tau}$ applied to all classes simultaneously, mapping the realistic, system-wide trade-offs enforced by a coherent decision rule.
It follows that the ROC cloud maps out the true, constrained performance envelope of the classifier in a multiclass setting.

In addition, the simplex decision rule is more nuanced than the simple thresholding of the OvR approach: it defines decision regions not with a simple cut, but with relational hyperplanes within the simplex. This allows for a more detailed exploration of the trade-offs between classes. 
In cases where the optimal decision boundary is relational (e.g., ``predict class A when its probability is high, especially if class B's is low''), this finer rule can discover efficiencies that the simpler OvR method cannot. This is why, for certain data distributions, operating points from the simplex method can locally exceed the standard ROC curve (see top row of \ref{fig:roc_clouds_combined}).

Finally, we introduce the DFP metric to summarize the distribution of the ROC points and get a measure of class separability.

\section{Conclusions}\label{sec:conclusions}

We introduced a threshold-based framework for multiclass classification that treats softmax outputs as points on the simplex and replaces the standard \textit{argmax} with a multidimensional threshold. 
This yields an inference-time decision rule that requires no retraining or post-hoc calibrators and supports \textit{a posteriori} optimization of any macro-aggregated score directly on validation data. 
Across several networks and datasets, we have seen from experiments that multidimensional threshold tuning can improve upon the \textit{argmax} baseline, with the largest gains appearing in unbalanced settings.

Also, we introduced \emph{ROC clouds} based on the simplex decision rule. These consist of the attainable $(\mathrm{FPR},\mathrm{TPR})$ operating points found by varying thresholds across the multidimensional simplex. 
In contrast to standard One-vs-Rest ROC curves, which sweep class-wise thresholds independently and thus trace an idealized class-isolated frontier, the ROC clouds capture the joint trade-offs enforced by one single natively multiclass decision rule. 
We further summarized each cloud with the \emph{Distance From Point} (DFP), the mean $L_1$ distance to $(0,1)$.

Finally, we point out that while multidimensional threshold tuning is clearly fully parallelizable, the optimization procedure becomes expensive if \textit{many} classes are involved due to the curse of dimensionality. In order to overcome this issue, one may look for the optimal threshold on a more \textit{rough} discretization of the simplex, sampling a fixed number of threshold values by means of a Monte Carlo approach or exploring the threshold space by means of suitable parameter searches (e.g., Bayesian schemes). 

\section*{Acknowledgments}
E.L. was supported by the HORIZON Europe ARCAFF Project, Grant No. 101082164. All authors acknowledge the Gruppo Nazionale per il Calcolo Scientifico - Istituto Nazionale di Alta Matematica (GNCS - INdAM).

\bibliographystyle{plain}
\bibliography{references}

@book{Vapnik1998,
  author = {Vapnik, Vladimir N.},
  publisher = {Wiley-Interscience},
  title = {Statistical Learning Theory},
  year = 1998
}

@article{zhang2000neural,
  title={Neural networks for classification: a survey},
author={Zhang, G.P.},
  journal={IEEE Transactions on Systems, Man, and Cybernetics, Part C (Applications and Reviews)},
  volume={30},
  number={4},
  pages={451--462},
  year={2000},
  publisher={IEEE}
}

@incollection{bridle1990probabilistic,
  title={Probabilistic interpretation of feedforward classification network outputs, with relationships to statistical pattern recognition},
  author={Bridle, John S},
  booktitle={Neurocomputing: Algorithms, architectures and applications},
  pages={227--236},
  publisher={Springer},
year = 1990
}

@article{tang2019novel,
  title={A novel perspective on multiclass classification: Regular simplex support vector machine},
  author={Tang, Long and Tian, Yingjie and Pardalos, Panos M},
  journal={Information Sciences},
  volume={480},
  pages={324--338},
  year={2019},
  publisher={Elsevier}
}

@inproceedings{martins2016softmax,
  title={From softmax to sparsemax: A sparse model of attention and multi-label classification},
  author={Martins, Andre and Astudillo, Ramon},
  booktitle={International conference on machine learning},
  pages={1614--1623},
  year={2016},
  organization={PMLR}
}

@inproceedings{medmnistv1,
    title = {MedMNIST Classification Decathlon: A Lightweight AutoML Benchmark for Medical Image Analysis},
    author = {Yang, Jiancheng and Shi, Rui and Ni, Bingbing},
    booktitle = {IEEE 18th International Symposium on Biomedical Imaging (ISBI)},
    pages = {191--195},
    year = {2021}
}

@article{medmnistv2,
  title={MedMNIST v2-A large-scale lightweight benchmark for 2D and 3D biomedical image classification},
  author={Yang, Jiancheng and Shi, Rui and Wei, Donglai and Liu, Zequan and Zhao, Lin and Ke, Bilian and Pfister, Hanspeter and Ni, Bingbing},
  journal={Scientific Data},
  volume={10},
  number={1},
  pages={41},
  year={2023},
  publisher={Nature Publishing Group UK London}
}

@article{pathmnist,
    title = {Predicting survival from colorectal cancer histology slides using deep learning: A retrospective multicenter study},
    author = {Kather, Jakob Nikolas AND Krisam, Johannes AND others},
    journal = {PLOS Medicine},
    publisher = {Public Library of Science},
    year = {2019},
    month = {01},
    volume = {16},
    pages = {1-22},
    number = {1},
}

@article{octmnist,
    title = {Identifying Medical Diagnoses and Treatable Diseases by Image-Based Deep Learning},
    author = {Daniel S. Kermany and Michael Goldbaum and others},
    journal = {Cell},
    volume = {172},
    number = {5},
    pages = {1122 - 1131.e9},
    year = {2018}
}

@article{Marchetti22,
title = {Score-{O}riented {L}oss ({SOL}) functions},
journal = {Pattern Recognition},
volume = {132},
pages = {108913},
year = {2022},
issn = {0031-3203},
doi = {https://doi.org/10.1016/j.patcog.2022.108913},
url = {https://www.sciencedirect.com/science/article/pii/S0031320322003946},
author = {F. Marchetti and S. Guastavino and M. Piana and C. Campi}
}

@article{guastavino2022implementation,
  title={Implementation paradigm for supervised flare forecasting studies: A deep learning application with video data},
  author={Guastavino, Sabrina and Marchetti, Francesco and Benvenuto, Federico and Campi, Cristina and Piana, Michele},
  journal={Astronomy \& Astrophysics},
  volume={662},
  pages={A105},
  year={2022},
  publisher={EDP Sciences}
}

@article{heese2023calibrated,
  title={Calibrated simplex-mapping classification},
  author={Heese, Raoul and Schmid, Jochen and Walczak, Micha{\l} and Bortz, Michael},
  journal={PLoS One},
  volume={18},
  number={1},
  pages={e0279876},
  year={2023},
  publisher={Public Library of Science San Francisco, CA USA}
}

@inproceedings{zadrozny2002transforming,
  title={Transforming classifier scores into accurate multiclass probability estimates},
  author={Zadrozny, Bianca and Elkan, Charles},
  booktitle={Proceedings of the eighth ACM SIGKDD international conference on Knowledge discovery and data mining},
  pages={694--699},
  year={2002}
}

@article{murphy2018machine,
  title={Machine learning: A probabilistic perspective (adaptive computation and machine learning series)},
  author={Murphy, Kevin P},
  journal={The MIT Press: London, UK},
  year={2018}
}

@article{xiao2017fashion,
  title={Fashion-mnist: a novel image dataset for benchmarking machine learning algorithms},
  author={Xiao, Han and Rasul, Kashif and Vollgraf, Roland},
  journal={arXiv preprint arXiv:1708.07747},
  year={2017}
}

@article{soloff2024building,
  title={Building a stable classifier with the inflated argmax},
  author={Soloff, Jake and Barber, Rina and Willett, Rebecca},
  journal={Advances in Neural Information Processing Systems},
  volume={37},
  pages={70349--70380},
  year={2024}
}

@article{mroueh2012multiclass,
  title={Multiclass learning with simplex coding},
  author={Mroueh, Youssef and Poggio, Tomaso and Rosasco, Lorenzo and Slotine, Jean-Jeacques},
  journal={Advances in Neural Information Processing Systems},
  volume={25},
  year={2012}
}

@article{roberts2023geometry,
  title={Geometry-aware adaptation for pretrained models},
  author={Roberts, Nicholas and Li, Xintong and Adila, Dyah and Cromp, Sonia and Huang, Tzu-Heng and Zhao, Jitian and Sala, Frederic},
  journal={Advances in Neural Information Processing Systems},
  volume={36},
  pages={49628--49658},
  year={2023}
}

@book {Goodfellow16,
    AUTHOR = {Goodfellow, Ian and Bengio, Yoshua and Courville, Aaron},
     TITLE = {Deep learning},
    SERIES = {Adaptive Computation and Machine Learning},
 PUBLISHER = {MIT Press, Cambridge, MA},
      YEAR = {2016},
     PAGES = {xxii+775},
      ISBN = {978-0-262-03561-3},
   MRCLASS = {68-02 (62-02 62H25 62H99 68T05)},
  MRNUMBER = {3617773},
}

@book{Harris13,
author = {Harris, David and Harris, Sarah},
title = {Digital Design and Computer Architecture, Second Edition},
year = {2012},
isbn = {0123944244},
publisher = {Morgan Kaufmann Publishers Inc.},
address = {San Francisco, CA, USA},
edition = {2nd},
}

@article{fang2019deep,
  title={Deep learning for automatic recognition of magnetic type in sunspot groups},
  author={Fang, Yuanhui and Cui, Yanmei and Ao, Xianzhi},
  journal={Advances in Astronomy},
  volume={2019},
  number={1},
  pages={9196234},
  year={2019},
  publisher={Wiley Online Library}
}

@book{pesnell2012solar,
  title={The solar dynamics observatory (SDO)},
  author={Pesnell, WD and Thompson, BJ and Chamberlin, PC},
  year={2012},
  publisher={Springer}
}

@article{scherrer2012helioseismic,
  title={The helioseismic and magnetic imager (HMI) investigation for the solar dynamics observatory (SDO)},
  author={Scherrer, Philip Hanby and Schou, Jesper and Bush, RI and Kosovichev, AG and Bogart, RS and Hoeksema, JT and Liu, Y and Duvall, TL and Zhao, J and Title, AM and others},
  journal={Solar Physics},
  volume={275},
  pages={207--227},
  year={2012},
  publisher={Springer}
}

@article{legnaro2025deep,
  title={Deep Learning for Active Region Classification: A Systematic Study from Convolutional Neural Networks to Vision Transformers},
  author={Legnaro, Edoardo and Guastavino, Sabrina and Piana, Michele and Massone, Anna Maria},
  journal={The Astrophysical Journal},
  volume={981},
  number={2},
  pages={157},
  year={2025},
  publisher={IOP Publishing}
}

@inproceedings{touvron2021training,
  title={Training data-efficient image transformers \& distillation through attention},
  author={Touvron, Hugo and Cord, Matthieu and Douze, Matthijs and Massa, Francisco and Sablayrolles, Alexandre and J{\'e}gou, Herv{\'e}},
  booktitle={International conference on machine learning},
  pages={10347--10357},
  year={2021},
  organization={PMLR}
}

@misc{rw2019timm,
  author       = {Ross Wightman},
  title        = {PyTorch Image Models},
  year         = {2019},
  publisher    = {GitHub},
  journal      = {GitHub repository},
  howpublished = {\url{https://github.com/huggingface/pytorch-image-models}},
  doi          = {10.5281/zenodo.4414861}
}

@article{fawcett2006introduction,
  title={An introduction to ROC analysis},
  author={Fawcett, Tom},
  journal={Pattern recognition letters},
  volume={27},
  number={8},
  pages={861--874},
  year={2006},
  publisher={Elsevier}
}

@article{bradley1997use,
  title={The use of the area under the ROC curve in the evaluation of machine learning algorithms},
  author={Bradley, Andrew P},
  journal={Pattern recognition},
  volume={30},
  number={7},
  pages={1145--1159},
  year={1997},
  publisher={Elsevier}
}

@article{hand2001simple,
  title={A simple generalisation of the area under the ROC curve for multiple class classification problems},
  author={Hand, David J and Till, Robert J},
  journal={Machine learning},
  volume={45},
  number={2},
  pages={171--186},
  year={2001},
  publisher={Springer}
}

@inproceedings{guo2017calibration,
  title={On calibration of modern neural networks},
  author={Guo, Chuan and Pleiss, Geoff and Sun, Yu and Weinberger, Kilian Q},
  booktitle={International conference on machine learning},
  pages={1321--1330},
  year={2017},
  organization={PMLR}
}

@inproceedings{kleiman2019aucmu,
  title={Auc$\mu$: A performance metric for multi-class machine learning models},
  author={Kleiman, Ross and Page, David},
  booktitle={International Conference on Machine Learning},
  pages={3439--3447},
  year={2019},
  organization={PMLR}
}

@article{allwein2000reducing,
  title={Reducing multiclass to binary: A unifying approach for margin classifiers},
  author={Allwein, Erin L and Schapire, Robert E and Singer, Yoram},
  journal={Journal of machine learning research},
  volume={1},
  number={Dec},
  pages={113--141},
  year={2000}
}

@article{wandishin2009multiclass,
  title={Multiclass ROC analysis},
  author={Wandishin, Matthew S and Mullen, Steven J},
  journal={Weather and Forecasting},
  volume={24},
  number={2},
  pages={530--547},
  year={2009}
}

@inproceedings{holmes2002multiclass,
  title={Multiclass alternating decision trees},
  author={Holmes, Geoffrey and Pfahringer, Bernhard and Kirkby, Richard and Frank, Eibe and Hall, Mark},
  booktitle={European Conference on Machine Learning},
  pages={161--172},
  year={2002},
  organization={Springer}
}

@inproceedings{ferri2003volume,
  title={Volume under the ROC surface for multi-class problems},
  author={Ferri, C{\'e}sar and Hern{\'a}ndez-Orallo, Jos{\'e} and Salido, Miguel Angel},
  booktitle={European conference on machine learning},
  pages={108--120},
  year={2003},
  organization={Springer}
}

@article{landgrebe2008efficient,
  title={Efficient multiclass ROC approximation by decomposition via confusion matrix perturbation analysis},
  author={Landgrebe, Thomas CW and Duin, Robert PW},
  journal={IEEE transactions on pattern analysis and machine intelligence},
  volume={30},
  number={5},
  pages={810--822},
  year={2008},
  publisher={IEEE}
}

\end{document}